\documentclass[10pt,twocolumn,letterpaper]{article}
\usepackage{cvpr}
\usepackage{graphicx} \graphicspath{{figures/}}
\usepackage{amsmath,amssymb,mathabx,mathtools,amsthm,nicefrac}
\usepackage{acronym}
\definecolor{cvprblue}{rgb}{0.21,0.49,0.74}\usepackage[pagebackref,breaklinks,colorlinks,allcolors=cvprblue]{hyperref}
\usepackage{balance}
\usepackage{xspace}
\usepackage{setspace}
\usepackage[skip=3pt,font=small]{caption}
\usepackage[dvipsnames,svgnames,x11names,table]{xcolor}
\usepackage[capitalize]{cleveref}
\usepackage{booktabs,tabularx,colortbl,multirow,array,makecell,tabularray}
\usepackage{enumitem}
\usepackage[misc]{ifsym}
\usepackage{siunitx}
\usepackage{pifont}

\makeatletter
\DeclareRobustCommand\onedot{\futurelet\@let@token\@onedot}
\def\@onedot{\ifx\@let@token.\else.\null\fi\xspace}
\def\eg{\emph{e.g}\onedot} 

\def\ie{\emph{i.e}\onedot}

\def\vs{\emph{vs}\onedot}

\def\etal{\emph{et al}\onedot}
\makeatother

\definecolor{mygray}{gray}{.9}

\newcommand{\mytablecolsep}{3pt}
\newcommand{\pub}[1]{{\color{gray}{\tiny{[{#1}]}}}}
\newcommand{\cmark}{\textcolor{green!70!black}{\ding{51}}} 
\newcommand{\xmark}{\textcolor{red!70!black}{\ding{55}}}   

\makeatletter
\renewcommand{\paragraph}{%
  \@startsection{paragraph}{4}%
  {\z@}{0ex \@plus 0ex \@minus 0ex}{-1em}%
  {\hskip\parindent\normalfont\normalsize\bfseries}%
}
\makeatother

\def\model{\textbf{\textsc{MEDVCR}}\xspace} 
\acrodef{sota}[SOTA]{state-of-the-art}
\acrodef{cg}[CG]{Counterfactual Generator}
\acrodef{crl}[CRL]{Counterfactual Representation Learning}
\acrodef{ddp}[DDP]{Dual Diagnostic Prediction}
\acrodef{AUC}[AUC]{Area Under the Receiver Operating Characteristic curve}
\acrodef{AP}[AP]{Average Precision}
\newcommand{\clinrule}[1]{\smallskip\noindent\textit{#1}\enspace}
\def\paperID{*****}
\def\confName{CVPR}
\def\confYear{2026}

\title{Clinically-Grounded Counterfactual Reasoning for Medical Video Diagnosis\vspace{-18pt}}

\author{
    Jianzhe Gao\textsuperscript{2} \quad
    Churan Wang\textsuperscript{1}$^{\,\textrm{\Letter}}$ \quad
    Weiyi Zhang\textsuperscript{3}$^{\,\textrm{\Letter}}$ \quad
    Jianghua Li\textsuperscript{3} \quad
    Li-An Li\textsuperscript{3} \vspace{3pt}\\
    Wenguan Wang\textsuperscript{2}$^{\,\textrm{\Letter}}$ \quad
    Yixin Zhu\textsuperscript{5,7,8,9} \quad
    Yizhou Wang\textsuperscript{4,6,7} \vspace{1pt}\\
    \small \href{https://gaozzzz.github.io/MedVCR/}{https://gaozzzz.github.io/MedVCR/} \vspace{-1pt}\\%
    \small\textsuperscript{1} Center for Data Science in Clinical Medicine, Peking University Third Hospital \vspace{-1pt}\\
    \small\textsuperscript{2} The State Key Lab of Brain-Machine Intelligence, Zhejiang University \vspace{-1pt}\\
    \small\textsuperscript{3} Department of Gynecology and Obstetrics, 7th Medical Center of Chinese PLA General Hospital \vspace{-1pt}\\
    \small\textsuperscript{4} School of Computer Science, Peking University \quad
    \small\textsuperscript{5} School of Psychological and Cognitive Sciences, Peking University \vspace{-1pt}\\
    \small\textsuperscript{6} State Key Lab of General AI, Peking University \quad
    \small\textsuperscript{7} Nat'l Eng. Research Center of Visual Technology \vspace{-1pt}\\\
    \small\textsuperscript{8} Beijing Key Laboratory of Behavior and Mental Health, Peking University \vspace{-1pt}\\
    \small\textsuperscript{9} Embodied Intelligence Lab, PKU-Wuhan Institute for Artificial Intelligence \vspace{-9pt}\\
}%

\begin{document}
\maketitle

\begin{abstract}
Clinical video diagnosis, in which physicians assess dynamic tissue responses across procedural stages, is critical for detecting diseases such as cervical and colorectal cancers. Recent spatiotemporal models map visual progressions directly to diagnostic outputs, yet overlook two hallmarks of expert reasoning: clinically grounded diagnostic principles and hypothesis-driven counterfactual thinking. This gap leads existing methods to conflate causal pathological cues with non-pathological variations, thereby limiting their reliability in data-scarce settings. Here we show that explicitly modeling counterfactual tissue evolution, guided by clinical rules that encode expert diagnostic principles, substantially improves diagnostic robustness---emulating the hypothesis-driven reasoning clinicians employ in practice. We introduce \model, comprising a \ac{cg} that synthesizes hypothetical tissue transitions via diffusion modeling, a \ac{crl} module that enforces temporal consistency, pathological separability, and counterfactual alignment, and a \ac{ddp} strategy that combines video-level context with frame-level counterfactual contrast. On two representative tasks, \model achieves 93.0\% Recall@1 on colposcopy (+10.2\%) and 94.8\% \ac{AP} on colonoscopy (+2.6\%), outperforming all \ac{sota} baselines. We anticipate that this counterfactual reasoning paradigm will open new avenues for vision-based clinical diagnostic tasks toward more transparent and interpretable decision support.
\end{abstract}

\begin{figure}[t!]
	\centering
    \small
    \includegraphics[width=\linewidth]{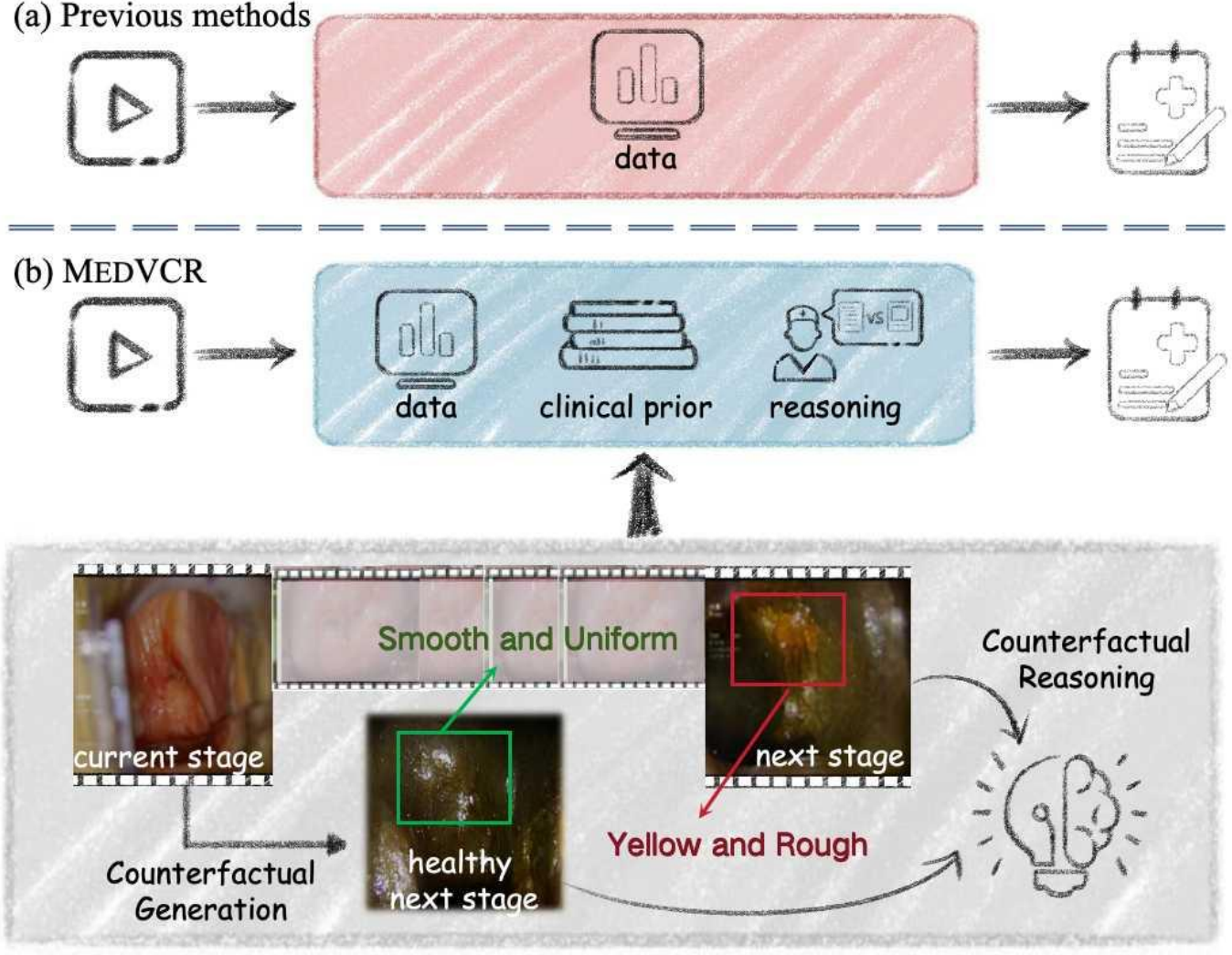}
	\caption{\textbf{Motivation and overview of \model.} (a) Previous methods fall into the data-driven paradigm that directly maps observations to diagnostic outputs. (b) The proposed \model incorporates clinical rules and models tissue evolution under different disease states. It performs counterfactual reasoning by contrasting factual observations with hypothetical alternatives to support reliable diagnosis.}
	\label{fig:introduce}
\end{figure}

\section{Introduction}

Clinical diagnosis in many diseases relies on video-level examinations, where physicians assess how tissue appearance evolves across procedural stages to determine underlying pathology~\cite{hall2023benefits,schreiberhuber2024cervical,joshi2025randomised}. This diagnostic process is often time-consuming, highly subjective, and operator-dependent, making automated medical video diagnosis increasingly valuable in clinical practice~\cite{liu2023skit,chen2024ultrasound}. Recent advances in video understanding have enabled end-to-end learning for medical video analysis~\cite{hu2023nurvid,pan2024mambasci,hong2025motionbench}, with mainstream methods leveraging convolution- or self-attention-based visual encoders~\cite{yang2024vivim,stidham2024using} to achieve promising performance in colonoscopy polyp detection~\cite{ma2021ldpolypvideo,choudhuri2025polypsegtrack}, laparoscopic surgery phase recognition~\cite{czempiel2021opera,gao2021trans,liu2023skit}, and fetal ultrasound standard plane localization~\cite{guo2024mmsummary,chen2024ultrasound,mishra2025mcat}.

Despite these advances, three fundamental challenges remain unresolved. \textbf{(i) Misinterpreting pathological evolution.} Prevailing models treat medical videos as sequences of visual changes~\cite{wang2023foundation,yang2024surgformer}, analyzing pixel-level variations while overlooking how tissues evolve across examination stages (\eg, the transient reactions to acetic acid or iodine). \textbf{(ii) Ignoring clinical diagnostic principles.} Relying purely on data-driven optimization without embedding explicit clinical knowledge~\cite{xie2021survey,huang2023self}, most existing methods are prone to conflating meaningful diagnostic cues with non-pathological variations from illumination shifts, reagent color variations, or camera motion. \textbf{(iii) Lacking hypothesis-driven reasoning.} Rather than distinguishing causal pathological cues from incidental correlations~\cite{castro2020causality,wang2021bilateral}, current models simply correlate observed patterns with diagnostic outputs. This stands in contrast to clinicians, who mentally simulate ``What if this tissue were benign instead of malignant?'' and compare hypothetical scenarios against observed reality---a counterfactual reasoning capacity, essential yet absent in most automated systems.

This hypothesis-driven reasoning is especially critical when learning from limited clinical cases~\cite{castro2020causality}. In practice, clinicians routinely simulate alternative pathological scenarios and contrast these hypothetical evolutions against actual tissue progressions~\cite{richens2020improving}. Through such counterfactual thinking, they extract generalizable diagnostic cues from sparse examples and separate true pathological signals from confounding factors---a capability that current automated methods notably lack. Addressing these challenges calls for a unified framework that simultaneously models pathology-conditioned tissue evolution, encodes clinical diagnostic principles as explicit constraints, and performs contrastive reasoning over factual and hypothetical observations---mirroring the cognitive process clinicians employ in practice.

To address these challenges, we propose \model, a hypothesis-driven counterfactual reasoning framework for medical video diagnosis that emulates clinical diagnostic thinking. As illustrated in \cref{fig:introduce}, \model integrates factual observations with clinically plausible hypothetical scenarios to simulate real-world diagnostic reasoning. It consists of three components. A \ac{cg} (\cref{sec:cg}) synthesizes realistic tissue transitions across clinical stages under specified pathological states via diffusion-based modeling. A \ac{crl} module (\cref{sec:crl}) learns spatiotemporal representations regularized by three clinical rules: \textit{temporal consistency} across examination stages, \textit{pathological separability} between disease patterns, and \textit{counterfactual alignment} with plausible pathological explanations. A \ac{ddp} strategy (\cref{sec:ddp}) aggregates global temporal dynamics while contrasting factual observations against counterfactual scenarios to yield reliable and interpretable predictions.

\model is evaluated under two representative settings. In fully supervised colposcopy biopsy-site localization, \model achieves 93.0\% Recall@1, exceeding the best prior method by 10.2\%. In weakly supervised colonoscopy polyp-frame detection, it attains 94.8\% \ac{AP}, surpassing the strongest baseline by 2.6\%. Extensive ablation studies further validate each component's contribution.

\section{Related Work}

\paragraph*{Medical video diagnosis}
Early methods~\cite{jin2017sv,twinanda2016endonet,twinanda2018rsdnet} apply frame-level CNNs that analyze each frame independently, neglecting the temporal evolution of tissue responses throughout examination. With the emergence of spatiotemporal backbones such as I3D~\cite{carreira2017quo}, SlowFast~\cite{feichtenhofer2019slowfast}, and ViViT~\cite{arnab2021vivit}, the field shifted toward end-to-end models that jointly capture spatial appearance and temporal progression. These backbones were first adopted for surgical phase recognition~\cite{jin2017sv,zisimopoulos2018deepphase,czempiel2021opera,ding2022exploring}, then extended to finer-grained surgical video analysis incorporating tool usage cues and instrument--tissue interactions~\cite{jin2020multi,tao2023last,chen2023surgical,rivoir2024pitfalls}. Parallel lines of work address medical video segmentation for precise lesion boundary delineation~\cite{deng2024memsam,yang2024vivim,chen2025stddnet,li2025semi,choudhuri2025polypsegtrack} and weakly supervised video detection that identifies pathological frames using only video-level labels~\cite{tian2022contrastive,madan2023self,mishra2025tier}. More recently, foundation models have extended the scope to high-complexity data such as whole-slide pathology images~\cite{xu2024whole} and multimodal clinical contexts~\cite{zhang2025multimodal}. Despite these advances, these methods focus primarily on observable visual variations and lack counterfactual reasoning---contradicting the clinical principle of determining pathology by mentally comparing observed tissue responses with hypothetical alternatives. \model addresses this gap by explicitly simulating and contrasting tissue evolution under different pathological conditions.

\paragraph*{Counterfactual reasoning}
Counterfactual reasoning, a cornerstone of causal inference and decision-making, enables models to explore hypothetical scenarios by predicting alternative outcomes under modified conditions~\cite{kim2023grounding,xiao2023masked,jeanneret2023adversarial,augustin2024dig,song2024doubly,wang2025omnidrive,shen2025image}. Early approaches integrate Structural Causal Models with deep generative architectures, leveraging normalizing flows and variational inference to infer exogenous noise under the no-unobserved-confounders assumption~\cite{pawlowski2020deep,yang2021causalvae,javaloy2023causal,melistas2024benchmarking}. GAN-based methods perform counterfactual inference through adversarial objectives or reparameterized attribute distributions to model interventional effects on image semantics~\cite{choi2018stargan,goetschalckx2019ganalyze,zhu2020cookgan,wang2021bilateral}, while VAE-based frameworks learn structured latent spaces that capture causal dependencies among high-level factors~\cite{yang2020causalvae,brehmer2022weakly}. More recently, diffusion models learn causal paths between semantic attributes or scene configurations by formulating noise abduction as a forward diffusion process~\cite{ma2024diffpo,zhang2024causaldiff,lin2025causal}. Recent work has adapted these ideas to medical image analysis, using counterfactual generation to enhance diagnostic interpretability, data diversity, and model reliability~\cite{yeganeh2025latent}. \model extends this paradigm to medical video diagnosis, enabling counterfactual reasoning over temporal tissue transitions across examination stages.

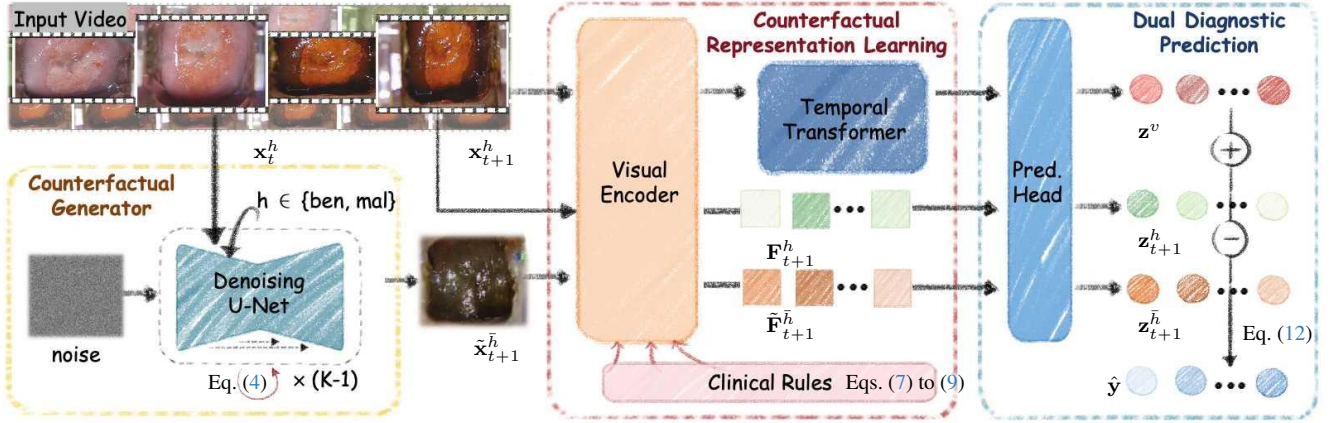
\begin{figure*}[t!]
    \centering
    \small
    \includegraphics[width=\linewidth]{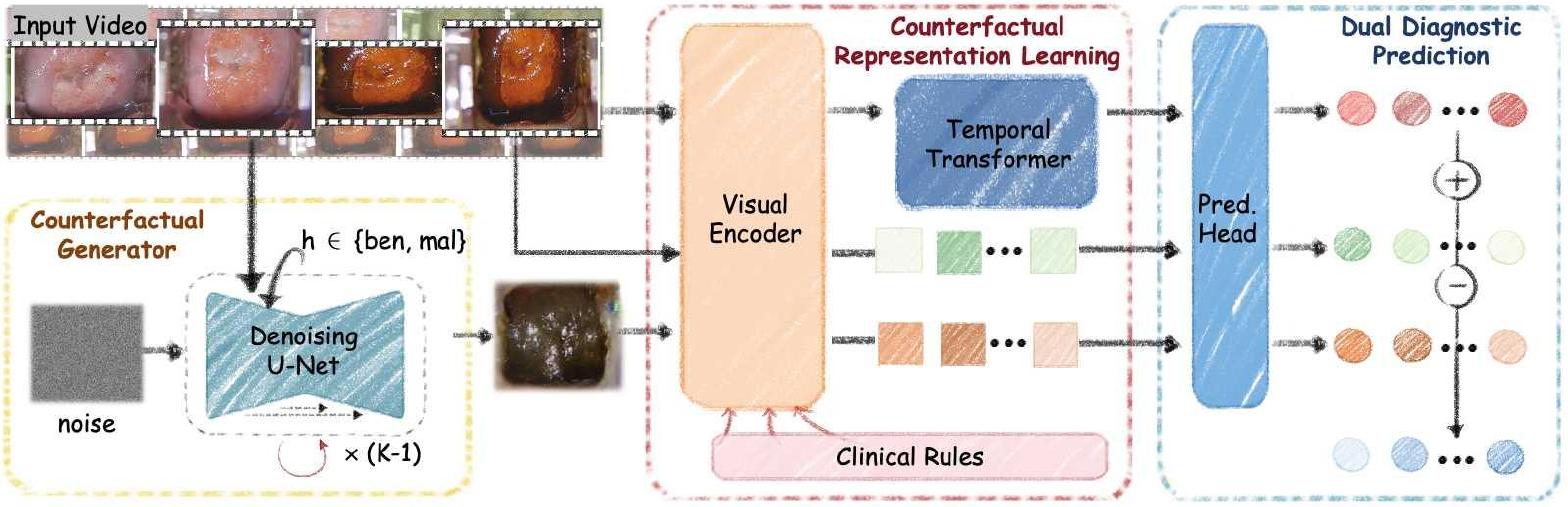}
    \put(-420,13.5){\footnotesize{\cref{eq:k}}}
    \put(-320,26.5){\footnotesize{$\tilde{\mathbf{x}}_{t+1}^{\bar{h}}$}}
    \put(-322,100.5){\footnotesize{$\mathbf{x}_{t+1}^{h}$}}
    \put(-403,100.5){\footnotesize{$\mathbf{x}_{t}^{h}$}}
    \put(-180,13.5){\footnotesize{\cref{eq:r1,eq:r2,eq:r3}}}
    \put(-30.5,32){\footnotesize{\cref{eq:y}}}
    \put(-70,35){\footnotesize{$\mathbf{z}_{t+1}^{\bar{h}}$}}
    \put(-70,67){\footnotesize{$\mathbf{z}_{t+1}^{h}$}}
    \put(-70,108){\footnotesize{$\mathbf{z}^{v}$}}
    \put(-81.5,12.5){\footnotesize{$\hat{\mathbf{y}}$}}
    \put(-210,35){\footnotesize{$\tilde{\mathbf{F}}_{t+1}^{\bar{h}}$}}
    \put(-210,62.5){\footnotesize{$\mathbf{F}_{t+1}^{h}$}}
    \caption{\textbf{Overview of \model.} Given a medical video sequence, the \acs{cg} (\cref{sec:cg}) synthesizes alternative tissue transitions under benign/malignant hypotheses. The \acs{crl} module (\cref{sec:crl}) encodes factual and counterfactual frames, enforcing clinical rules for temporal consistency, pathological separability, and counterfactual alignment. The \acs{ddp} strategy (\cref{sec:ddp}) then integrates temporal context with frame-level counterfactual contrast to produce diagnoses.}
    \label{fig:framework}
\end{figure*}

\section{Method}

\paragraph{Problem formulation} Medical video analysis aims to infer diagnostic cues that reflect the underlying pathological condition and its temporal evolution. Given a video sequence $\mathcal{V}=\{\mathbf{x}_t\}_{t=1}^{T}$ recording progressive tissue changes across examination stages, each frame $\mathbf{x}_t$ is influenced by its stage $s_t$, the latent health state $h\!\in\!\{\text{benign},\text{malignant}\}$, and nuisance factors such as illumination or camera motion. The objective is to produce clinically meaningful predictions $\hat{\mathbf{y}}\!\in\![0,1]^{P}$, where $P$ denotes the number of prediction points that highlight diagnostically significant regions or potential sampling sites.

\paragraph{Overall design (\cref{fig:framework})} \model integrates counterfactual reasoning with medical video diagnosis to enhance diagnostic reliability. Specifically, a pretrained \ac{cg} (\cref{sec:cg}) synthesizes alternative tissue transitions that serve as counterfactual supervision. Through \ac{crl} (\cref{sec:crl}), spatiotemporal representations are captured and regularized by three clinical rules. The learned representations are then fed into a \ac{ddp} strategy (\cref{sec:ddp}) that integrates global temporal context with localized counterfactual evidence for final diagnosis.

\subsection{Counterfactual Generator (CG)}\label{sec:cg}

Physicians diagnose not only by observing how tissue appearance evolves across examination stages, but also by mentally comparing the observed progression with alternative outcomes under different pathological states--an implicit counterfactual reasoning process. Inspired by this, \acs{cg} is designed to synthesize realistic tissue transitions conditioned on specified health conditions.

Formally, the generator $\mathcal{G}$ is formulated as a conditional diffusion model. Given a reference frame $\mathbf{x}_t$ from stage $s_t$ and a target health condition $h$, $\mathcal{G}$ generates the corresponding frame $\tilde{\mathbf{x}}_{t+1}^h$ at the subsequent stage $s_{t+1}$ by estimating the expected tissue appearance under the specified pathological state. The generative process comprises three components: a fixed \textit{forward diffusion} that corrupts real tissue frames, a learnable \textit{reverse denoising} that restores diagnostic information, and an iterative \textit{sampling process} that synthesizes tissue transitions conditioned on any health state.

\paragraph{Forward diffusion}
The forward process follows a fixed Markov chain that progressively corrupts the target frame $\mathbf{x}_{t+1}$ by adding Gaussian noise across $K$ diffusion steps:
{%
    \small%
    \begin{equation}\label{eq:fwd}
        q(\boldsymbol{\epsilon}_k \mid \mathbf{x}_{t+1}) = \mathcal{N}\!\Big(\boldsymbol{\epsilon}_k;\,\sqrt{\bar{\alpha}_k}\,\mathbf{x}_{t+1},\,(1-\bar{\alpha}_k)\mathbf{I}\Big),
    \end{equation}%
}%
where $k\!\in\!\{1,\!\dots,\!K\}$, $\bar{\alpha}_k\!=\!\!\prod_{i=1}^{k}\!\alpha_i$ denotes the cumulative noise schedule, and $\boldsymbol{\epsilon}_k$ represents the progressively noised frame at diffusion step $k$. As $k$ increases, the tissue structure and color cues are gradually replaced by random noise, yielding a fully stochastic signal at the final step $K$. This diffusion process defines the degradation trajectory that the reverse model later learns to invert.

\paragraph{Reverse denoising}
This process learns to iteratively recover the clean frame from its noisy counterpart through a parameterized conditional transition. At each diffusion step $k$, a U-shaped network $\mathcal{F}^u$ predicts the step-specific noise component $\hat{\boldsymbol{\epsilon}}_k$ from the current noisy input $\boldsymbol{\epsilon}_k$:
{%
    \small%
    \begin{equation}\label{eq:u}
        \hat{\boldsymbol{\epsilon}}_k = \mathcal{F}^{u}(\boldsymbol{\epsilon}_k,\mathbf{x}_t,h,k).
    \end{equation}%
}%
Here, both $\boldsymbol{\epsilon}_k$ and $\mathbf{x}_t$ are encoded by the same visual backbone to extract spatial context, $h$ is projected into a latent vector that modulates pathological manifestation, and $k$ is embedded via sinusoidal positional encoding to represent diffusion progress. Following the standard DDPM denoising formulation~\cite{sohl2015deep}, the predicted noise $\hat{\boldsymbol{\epsilon}}_k$ is removed to obtain a cleaner estimate at the previous step:
{%
    \small%
    \begin{equation}
        \boldsymbol{\epsilon}_{k-1} = \frac{1}{\sqrt{\alpha_k}} \left(\boldsymbol{\epsilon}_k - \frac{\beta_k}{\sqrt{1-\bar{\alpha}_k}}\hat{\boldsymbol{\epsilon}}_k\right),
    \end{equation}%
}%
where $\alpha_k$ and $\beta_k$ define the predefined noise schedule controlling the denoising rate. This process gradually restores spatial structures and chromatic patterns coherent with the specified pathological condition.

\paragraph{Sampling process}
Built upon the learned denoising process, the generator $\mathcal{G}$ synthesizes tissue appearances under different pathological conditions. Starting from Gaussian noise $\boldsymbol{\epsilon}_{k}\!\sim\!\mathcal{N}(0,\mathbf{I})$, $\mathcal{G}$ iteratively applies $\mathcal{F}^u$ across $k$ steps to reconstruct the tissue frame:
{%
    \small%
    \begin{equation}\label{eq:k}
        \tilde{\mathbf{x}}_{t+1}^{h} = \mathcal{G}(\mathbf{x}_t, h) = \mathcal{F}^{u,1:k}\!(\boldsymbol{\epsilon}_{k};\,\mathbf{x}_t,h),
    \end{equation}%
}%
where $\mathcal{F}^{u,1:k}$ denotes the sequential composition of all denoising steps from $k$ to $1$. This iterative process transforms random noise into a coherent tissue image whose structure and color patterns evolve consistently with the specified health condition $h$. Generating both benign and malignant variants enables the model to simulate alternative diagnostic trajectories for counterfactual reasoning.

\subsection{Counterfactual Representation Learning}\label{sec:crl}

Given the pretrained $\mathcal{G}$, the medical video learner is trained to extract spatiotemporal representations from both factual video sequences and counterfactual hypotheses. A set of clinical rules is further introduced to enforce expected invariances and separabilities across temporal and pathological dimensions, encouraging the learner to derive representations that are both physiologically coherent and diagnostically interpretable.

\paragraph{Medical video learner}
The learner consists of a visual encoder $\mathcal{F}^{e}$ and a temporal Transformer $\mathcal{F}^{t}$, forming a hierarchical architecture for spatiotemporal representation learning. Given a video sequence $\mathcal{V} = \{\mathbf{x}_t\}_{t=1}^{T}$, overlapping clips of length $L$ are extracted with a temporal stride to preserve continuity, yielding a set of clips $\{\mathcal{C}_i\}_{i=1}^{C}$, where each clip $\mathcal{C}_i = \{\mathbf{x}_t\}_{t=i}^{i+L-1}$. Each clip is processed by $\mathcal{F}^{e}$ (\ie, I3D~\cite{carreira2017quo}) to obtain frame-level embeddings:
{%
    \small%
    \begin{equation}\label{eq:enc}
        \mathbf{F}_{i}^{e} = \mathcal{F}^{e}(\mathcal{C}_i) \in \mathbb{R}^{L \times d}.
    \end{equation}%
}%
Here, each row of $\mathbf{F}_{i}^{e}$ corresponds to a frame representation and $d$ denotes the feature dimension. All $\mathbf{F}_{i}^{e}$ are concatenated into a unified temporal sequence $\mathbf{F}^{e}\!\in\!\mathbb{R}^{(C \cdot L)\times d}$. A learnable temporal token $\mathbf{F}^{\tau}\!\in\!\mathbb{R}^{d}$ is then prepended to $\mathbf{F}^{e}$, enabling the learner to aggregate stage-wise dynamics across the full sequence. The resulting sequence is fed into $\mathcal{F}^{t}$ to capture long-range dependencies:
{%
    \small%
    \begin{equation}\label{eq:v}
        \mathbf{F}^{v} = \mathcal{F}^{t}([\mathbf{F}^{\tau}, \mathbf{F}^{e}]) \in \mathbb{R}^{(C \cdot L + 1) \times d},
    \end{equation}%
}%
where $[\cdot, \cdot]$ denotes concatenation. The final representation $\mathbf{F}^{v}$ encodes both local frame-wise patterns and global progression dynamics across examination stages.

\paragraph{Clinical rules}
Physiological and diagnostic knowledge is incorporated into representation learning through three clinical rules. Each rule is formulated using frame pairs $(\mathbf{x}_t, \mathbf{x}_{t+1})$ and counterfactual frames $\tilde{\mathbf{x}}_{t+1}^{h}$. Specifically, the pairs $(\mathbf{x}_t, \mathbf{x}_{t+1})$ represent consecutive examination stages $(s_t, s_{t+1})$ of the same tissue region, and the counterfactual frame $\tilde{\mathbf{x}}_{t+1}^{h}$ is generated by $\mathcal{G}$ from $\mathbf{x}_t$ under health condition $h\!\in\!\{\text{benign},\text{malignant}\}$ to model alternative diagnostic outcomes. All frames are encoded by $\mathcal{F}^{e}$ to obtain representations $\{\mathbf{F}_{t}, \mathbf{F}_{t+1}, \tilde{\mathbf{F}}_{t+1}^{\text{ben}}, \tilde{\mathbf{F}}_{t+1}^{\text{mal}}\}$, where $\tilde{\mathbf{F}}$ denotes features extracted from counterfactual frames $\tilde{\mathbf{x}}$.

\clinrule{Rule 1: Temporal consistency.}
\textit{Pathological status remains stable throughout medical examination sequences. The same tissue region, whether benign or malignant, maintains its diagnostic identity across different examination stages, despite variations in appearance due to reagent reactions, illumination differences, or procedural manipulation.}

This rule implies that the diagnostic information encoded in features should remain consistent across examination stages for the same tissue region. In information-theoretic terms, the encoder is expected to preserve health-related information while suppressing correlations with stage-dependent factors:
{%
    \small%
    \begin{equation}\label{eq:r1}
        \mathcal{M}(\mathbf{F}_{t};\, h) \approx \mathcal{M}(\mathbf{F}_{t+1};\, h) \gg \mathcal{M}(\mathbf{F}_{t},\,\mathbf{F}_{t+1};\, s_{t}, s_{t+1}),
    \end{equation}%
}%
where $\mathcal{M}(X;Y)$ denotes the mutual information between random variables $X$ and $Y$, and $s_t$, $s_{t+1}$ denote the corresponding examination stages. This relation ensures that the representations retain diagnostic information while remaining invariant to stage-specific variations.

\clinrule{Rule 2: Pathological separability.}
\textit{Benign and malignant tissues arise from distinct biological processes and exhibit discernible morphological and chromatic signatures.}

This rule enforces a clear separation of pathological conditions in the latent representation space. From an information-theoretic perspective, the features should collectively provide strong diagnostic information while remaining independent across pathological categories:
{%
    \small%
    \begin{equation}\label{eq:r2}
        \mathcal{M}(h;\, \mathbf{F}_{t+1}^{\text{ben}}) + \mathcal{M}(h;\, \mathbf{F}_{t+1}^{\text{mal}}) \gg \mathcal{M}(\mathbf{F}_{t+1}^{\text{ben}};\, \mathbf{F}_{t+1}^{\text{mal}}).
    \end{equation}%
}%
The left-hand terms encourage each pathological representation to preserve strong diagnostic dependency on $h$, while the right-hand term penalizes mutual dependence between benign and malignant features, ensuring a well-separated and clinically interpretable latent space.

\clinrule{Rule 3: Counterfactual alignment.}
\textit{Clinical diagnosis relies on the concordance between observed tissue characteristics and the expected manifestations of the true pathological state, while maintaining clear divergence from those associated with alternative conditions.}

This rule requires the encoder to align factual observations with their pathology-consistent counterfactual counterparts while separating them from incompatible ones. Formally:
{%
    \small%
    \begin{equation}\label{eq:r3}
    \mathcal{M}(\mathbf{F}_{t+1}^{h};\, \tilde{\mathbf{F}}_{t+1}^{h}) \gg \mathcal{M}(\mathbf{F}_{t+1}^{h};\, \tilde{\mathbf{F}}_{t+1}^{\bar{h}}),
    \end{equation}%
}%
where $\bar{h}$ denotes the opposite health state. This relation enforces diagnostic alignment with the true pathology while suppressing agreement with incompatible assumptions, thereby promoting clinically coherent counterfactual reasoning in the learned representation space.

\subsection{Dual Diagnostic Prediction (DDP)}\label{sec:ddp}

Clinical diagnosis follows a hierarchical reasoning process: physicians evaluate the entire examination video to assess global tissue response patterns while simultaneously examining individual keyframes to identify localized pathological evidence. Motivated by this, \ac{ddp} strategy integrates video-level temporal context with frame-level counterfactual analysis for comprehensive diagnostic predictions.

\paragraph{Video-level assessment}
The entire examination sequence $\mathcal{V}$ is processed by the medical video learner to obtain the video-level feature $\mathbf{F}^{v}$ (\cref{eq:v}), which is subsequently passed through the prediction head $\mathcal{F}^{p}$ to yield logits $\mathbf{z}^{v}$:
{%
    \small%
    \begin{equation}\label{eq:zv}
        \mathbf{z}^{v} = \mathcal{F}^{p}\!\big(\mathbf{F}^{v}\big) \in \mathbb{R}^{P},
    \end{equation}%
}%
where $P$ is the number of diagnostic outcomes.

\paragraph{Frame-level analysis}
To capture localized pathological evidence, both the factual keyframe $\mathbf{x}_{t+1}^{h}$ and its counterfactual counterpart $\tilde{\mathbf{x}}_{t+1}^{\bar{h}}$ are independently analyzed to provide complementary diagnostic perspectives:
{%
    \small%
    \begin{equation}\label{eq:zf}
        \mathbf{z}_{t+1}^{h} = \mathcal{F}^{p}\!\big(\mathcal{F}^{e}(\mathbf{x}_{t+1}^{h})\big),
        \quad
        \mathbf{z}_{t+1}^{\bar{h}} = \mathcal{F}^{p}\!\big(\mathcal{F}^{e}(\tilde{\mathbf{x}}_{t+1}^{\bar{h}})\big),
    \end{equation}%
}%
where $\mathbf{z}_{t+1}^{h},\,\mathbf{z}_{t+1}^{\bar{h}} \in \mathbb{R}^{P}$ represent per-site diagnostic logits from the factual and counterfactual analyses, respectively.

\paragraph{Diagnostic fusion}
The final prediction integrates global temporal context with localized counterfactual contrast:
{%
    \small%
    \begin{equation}\label{eq:y}
        \hat{\mathbf{y}} = \mathcal{F}^\sigma\!\big(
        \mathbf{z}^{v} + \mathbf{z}_{t+1}^{h} - \mathbf{z}_{t+1}^{\bar{h}}
        \big)
        \in [0,1]^{P},
    \end{equation}%
}%
where $\mathcal{F}^\sigma(\cdot)$ denotes the element-wise sigmoid function, producing independent probabilities for each diagnostic outcome. This formulation combines video-level progression dynamics $\mathbf{z}^{v}$ with frame-level evidence $\mathbf{z}_{t+1}^{h}$ supporting the true pathology, while simultaneously suppressing cues aligned with the alternative hypothesis $\mathbf{z}_{t+1}^{\bar{h}}$, thereby realizing differential diagnostic reasoning at the prediction level.

\subsection{Hybrid Loss Function}

The generator $\mathcal{G}$ is first pretrained to produce counterfactual supervision. Subsequently, the medical video learner and diagnostic head are jointly optimized to ensure physiologically consistent and diagnostically faithful predictions.

\paragraph{Generator loss}
The generator $\mathcal{G}$ is trained to reconstruct pathology-consistent frames through a denoising diffusion process. At each diffusion step $k$, the U-shaped denoising network $\mathcal{F}^{u}$ predicts the noise component $\hat{\boldsymbol{\epsilon}}_k$ from the current noisy input $\boldsymbol{\epsilon}_k$, conditioned on $\mathbf{x}_t$ and $h\!\in\!\{\text{benign},\text{malignant}\}$. The generator objective minimizes the expected mean-squared discrepancy between the predicted and reference noise across diffusion timesteps:
{%
    \small%
    \begin{equation}\label{eq:lgen}
        \mathcal{L}^{\mathrm{gen}}
        =
        \mathbb{E}_{\mathbf{x}_t, \mathbf{x}_{t+1}, h, k}\!
        \Big[
        \|\boldsymbol{\epsilon}_k - \hat{\boldsymbol{\epsilon}}_k\|_2^2
        \Big],
    \end{equation}%
}%
where $\hat{\boldsymbol{\epsilon}}_k = \mathcal{F}^{u}(\boldsymbol{\epsilon}_k, \mathbf{x}_t, h, k)$ (\cref{eq:u}). This encourages $\mathcal{G}$ to synthesize temporally coherent tissue transitions under the specified pathological condition.

\paragraph{Visual encoder loss}
All clinical rule constraints are imposed on frame-level embeddings extracted by the visual encoder $\mathcal{F}^{e}$, ensuring physiologically consistent representations before temporal aggregation by $\mathcal{F}^{t}$.

To maintain diagnostic stability across consecutive stages (Rule~1), a temporal contrastive loss encourages representations from adjacent stages to remain consistent:
{%
    \small%
    \begin{equation}\label{eq:ltemp}
        \mathcal{L}^{\mathrm{temp}} = 1 - \mathcal{F}^{\mathrm{sim}}(\mathbf{F}_{t}^{h}, \mathbf{F}_{t+1}^{h}),
    \end{equation}%
}%
where $\mathcal{F}^{\mathrm{sim}}(\cdot,\cdot)$ denotes normalized cosine similarity.

To encourage discriminative encoding of pathological states (Rule~2), a soft separability loss guides representations to reflect diagnostic differences:
{%
    \small%
    \begin{equation}\label{eq:lsep}
        \mathcal{L}^{\mathrm{sep}} = \mathcal{F}^{\mathrm{sim}}(\mathbf{F}_{t+1}^{h}, \mathbf{F}_{t}^{h}) - \mathcal{F}^{\mathrm{sim}}(\mathbf{F}_{t+1}^{h}, \tilde{\mathbf{F}}_{t+1}^{\bar{h}}).
    \end{equation}%
}%

To ensure alignment between factual and counterfactual reasoning (Rule~3), a counterfactual alignment loss adopts a triplet formulation that pulls features toward their pathology-consistent counterparts while pushing them away from incompatible ones:
{%
    \small%
    \begin{equation}\label{eq:lalign}
        \mathcal{L}^{\mathrm{align}} = \max\!\big(0,\,m +
        \mathcal{F}^{\mathrm{sim}}(\mathbf{F}_{t+1}^{h},\, \tilde{\mathbf{F}}_{t+1}^{\bar{h}}) - \mathcal{F}^{\mathrm{sim}}(\mathbf{F}_{t+1}^{h},\, \tilde{\mathbf{F}}_{t+1}^{h})
        \big),
    \end{equation}%
}%
where $m$ denotes the margin governing alignment strength.

\paragraph{Diagnostic loss} The diagnostic head is optimized using a supervised objective that compares the predicted probabilities $\hat{\mathbf{y}}$ with the ground-truth labels $\mathbf{y} \in \{0,1\}^{P}$:
{%
    \small%
    \begin{equation}\label{eq:ldiag}
        \mathcal{L}^{\mathrm{diag}} = -\frac{1}{P}\sum_{k=1}^{P}\big[y_k \log \hat{y}_k + (1 - y_k)\log(1 - \hat{y}_k)\big].
    \end{equation}%
}%

\subsection{Implementation Details}

\paragraph{Training}
All experiments are implemented in PyTorch and trained on a single NVIDIA RTX~4090 GPU with batch size~8. The \acs{cg} adopts a U-Net backbone with diffusion timesteps $K{=}1000$, trained for 20k iterations using AdamW with a cosine learning rate schedule and a base learning rate of $10^{-4}$. The visual encoder $\mathcal{F}^e$ is initialized from pretrained I3D~\cite{carreira2017quo}. Images are resized to $256{\times}256$ across all tasks. Please refer to \cref{sec:td} for more details.

\paragraph{Inference}
During inference, the input video sequence $\mathcal{V}$ is divided into overlapping clips and encoded by $\mathcal{F}^e$ to obtain global spatiotemporal representations. For frame-level diagnostic reasoning, we follow standard clinical screening protocols where observed tissues are treated as suspected cases ($h{=}\text{mal}$) to enable falsification. Accordingly, the counterfactual target is set to benign ($\bar{h}{=}\text{ben}$). To obtain the counterfactual reference for the current frame $\mathbf{x}_t$, the generator $\mathcal{G}$ uses the preceding frame $\mathbf{x}_{t-1}$ to synthesize the benign estimate $\tilde{\mathbf{x}}_t^{\bar{h}}$. Finally, \acs{ddp} module computes the diagnostic probabilities $\hat{\mathbf{y}}$ by integrating the video-level temporal context with the contrastive features derived from the pair $(\mathbf{x}_t, \tilde{\mathbf{x}}_t^{\bar{h}})$.

\section{Experiments}

To demonstrate the versatility and effectiveness of \model, experiments are conducted under two representative supervision settings reflecting common clinical video analysis scenarios: \textbf{fully supervised learning} (\cref{sec:colp_va}) with detailed frame-level pathological annotations, and \textbf{weakly supervised learning} (\cref{sec:colon_va}) using only video-level diagnostic labels. Each setting addresses clinically significant tasks embodying typical challenges, while ablation studies (\cref{sec:ablation}) systematically analyze individual component contributions. Additional validation on medical image analysis is provided in \cref{sec:addition_exp}.

\subsection{Fully Supervised Medical Video Diagnosis}\label{sec:colp_va}

To evaluate \model under the fully supervised setting, biopsy-site localization in colposcopy videos is adopted as the representative task. Since biopsy sites lie on fixed clock-face positions, this task is formulated as a multi-label classification problem.

\paragraph{Task description}
Colposcopy is a standard gynecological procedure for cervical cancer prevention and early detection~\cite{xue2025deep}. During the examination, a colposcope equipped with a magnified optical system is positioned outside the vagina to observe the cervical epithelium under different reagents such as saline, acetic acid, and iodine solution~\cite{brisson2019global}. These reagents induce transient color and texture changes that help reveal abnormal epithelial patterns and potential precancerous lesions. The analysis of colposcopy videos focuses on identifying regions of clinical interest for biopsy.

\paragraph{Dataset}
The in-house dataset comprises 623 patient-specific colposcopy examination records, each containing a complete four-stage video (\ie, saline, acetic acid, alcohol, and iodine) along with the corresponding stage keyframes and pathology report. The dataset covers a wide spectrum of cervical conditions, including normal tissues, low-grade lesions, and high-grade intraepithelial neoplasia. Each case is independently reviewed by three senior gynecologists, who delineate biopsy sites and verify pathological outcomes based on histopathological reports.

\paragraph{Metrics}
Performance is evaluated using four metrics: Recall, Precision, Accuracy, and Tolerant Recall (Recall@1). Recall@1 considers a prediction correct if it falls within one adjacent position of the true biopsy site, reflecting the clinically acceptable localization tolerance.

\paragraph{Compared methods}
Comparisons are conducted using both general and medical video understanding models to extract spatiotemporal representations.
\begin{itemize}
    \item \textit{General methods.} Representative natural video understanding models, including TimeSformer~\cite{bertasius2021space}, VideoSwin~\cite{liu2022video}, and VideoMAEv2~\cite{wang2023videomae}, are used.
    \item \textit{Medical methods.} Since colposcopy-specific models are underexplored, typical video encoders from surgical (SurgFormer~\cite{yang2024surgformer} and VideoCutMix~\cite{dhanakshirur2024videocutmix}), endoscopic (SurgVLP~\cite{yuan2025learning}, PolypSegTrack~\cite{choudhuri2025polypsegtrack}, and STDDNet~\cite{chen2025stddnet}), and ultrasound (EchoPrime~\cite{nature2025}) domains are adopted for comparison.
    \item \textit{Ours.} The acetic acid and iodine stages are chosen as $s_t$ and $s_{t+1}$, consistent with clinical practice where physicians primarily rely on these two stages for assessment.
\end{itemize}

\begin{table}[t!]
    \centering
    \small
    \setlength\tabcolsep{\mytablecolsep}
    \caption{\textbf{Quantitative results of fully supervised colposcopy video analysis on our in-house dataset.} Results are obtained through five-fold cross-validation. Recall@1 considers predictions within one adjacent position as correct. See \cref{sec:colp_va} for details.}
    \label{table:colposcopy}
    \resizebox{\linewidth}{!}{%
        \begin{tabular}{cccccc}
            \toprule
            \multirow{1}{*}{Category} & \multirow{1}{*}{Methods} 
            & Recall$\uparrow$ & Precision$\uparrow$ & Acc.$\uparrow$ & Recall@1$\uparrow$ \\
            \midrule
            \multirow{3}{*}{\textit{General}} 
            & TimeSformer\pub{ICML21}~\cite{bertasius2021space} & 54.8 & 57.9 & 25.6 & 70.4 \\
            & VideoSwin\pub{CVPR22}~\cite{liu2022video}          & 60.9 & 61.8 & 29.7 & 75.2 \\
            & VideoMAEv2\pub{CVPR23}~\cite{wang2023videomae}     & 65.3 & 65.9 & 33.5 & 77.6 \\
            \midrule
            \multirow{6}{*}{\textit{Medical}} 
            & VideoCutMix\pub{MICCAI24}~\cite{dhanakshirur2024videocutmix}  & 67.3 & 68.2 & 39.8 & 81.5 \\
            & SurgFormer\pub{MICCAI24}~\cite{yang2024surgformer}            & 70.1 & 66.8 & 41.2 & 82.8 \\
            & PolypSegTrack\pub{MICCAI25}~\cite{choudhuri2025polypsegtrack} & 65.5 & 63.8 & 36.5 & 79.6 \\
            & EchoPrime\pub{Nature25}~\cite{nature2025}                    & 67.8 & 66.1 & 37.4 & 80.2 \\
            & STDDNet\pub{ICCV25}~\cite{chen2025stddnet}                    & 66.8 & 67.1 & 38.1 & 82.3 \\
            & SurgVLP\pub{MIA25}~\cite{yuan2025learning}                  & 68.9 & 65.9 & 38.7 & 82.0 \\
            \hline
            \multicolumn{2}{c}{\textbf{\textsc{MedVCR} (Ours)}} 
            & \textbf{80.3} & \textbf{74.4} & \textbf{55.0} & \textbf{93.0} \\
            \bottomrule
        \end{tabular}%
    }%
\end{table}

\paragraph{Quantitative results}
Results are reported in \cref{table:colposcopy}. (i)~Directly applying general video understanding models yields limited performance (\ie, 70.4\% Recall@1 for TimeSformer~\cite{bertasius2021space}), revealing a substantial domain gap between natural videos and clinical examinations. (ii)~Models fine-tuned on related medical domains achieve moderate gains (\ie, 82.8\% Recall@1 for SurgFormer~\cite{yang2024surgformer}), suggesting that medical priors can generalize across clinical tasks. (iii)~\model achieves the best overall performance with \textbf{93.0}\% Recall@1, demonstrating strong clinical applicability in accurately localizing biopsy sites.

\subsection{Weakly Supervised Medical Video Diagnosis}\label{sec:colon_va}

Weakly supervised medical video diagnosis involves learning from video-level diagnostic labels without frame-level supervision. In this setting, \model is evaluated on weakly supervised polyp-frame detection in colonoscopy.

\begin{table}[t!]
    \centering
    \small
    \setlength\tabcolsep{\mytablecolsep}
    \caption{\textbf{Quantitative results of weakly supervised colonoscopy video analysis on the combined Kvasir~\cite{borgli2020hyperkvasir} and LDPolypVideo~\cite{ma2021ldpolypvideo} dataset.} Results are obtained through five-fold cross-validation. See \cref{sec:colon_va} for details.}
    \label{tab:polyp_anomaly}
    \begin{tabular}{cccc}
        \toprule
        \multirow{1}{*}{Category} & \multirow{1}{*}{Methods} 
        & AP $\uparrow$ & AUC $\uparrow$ \\
        \midrule
        \multirow{5}{*}{\textit{General}} 
        & GCN-Ano\pub{ICCV19}~\cite{zhong2019graph} & 75.4 & 92.1 \\
        & CLAWS\pub{ECCV20}~\cite{zaheer2020claws}  & 80.4 & 95.6 \\
        & MIST\pub{CVPR21}~\cite{feng2021mist}      & 72.9 & 94.5 \\
        & RTFM\pub{ICCV21}~\cite{tian2021weakly}    & 78.0 & 96.3 \\
        & UR-DMU\pub{AAAI23}~\cite{zhou2023dual}    & 79.3 & 93.7 \\
        \midrule
        \multirow{4}{*}{\textit{Medical}} 
        & CTMIL\pub{MICCAI22}~\cite{tian2022contrastive} & 86.6 & 98.4 \\
        & Endo-FM\pub{MICCAI23}~\cite{wang2023foundation}      & 89.2 & 97.6 \\
        & EchoPrime\pub{Nature25}~\cite{nature2025}                   & 75.7 & 92.4 \\
        & Fadmb\pub{PR25}~\cite{luo2025fadmb}      & 92.2 & 99.4 \\
        \hline
        \multicolumn{2}{c}{\textbf{\textsc{MedVCR} (Ours)}} 
        & \textbf{94.8} & \textbf{99.6} \\
        \bottomrule
    \end{tabular}%
\end{table}

\paragraph{Task description}
Colonoscopy serves as the primary technique for the early detection and prevention of colorectal cancer~\cite{fan2020pranet}. In clinical practice, a flexible endoscope equipped with a camera is inserted through the rectum to observe the mucosal surface of the colon in real time~\cite{cao2024robotic}. The procedure focuses on detecting polyps---abnormal tissue growths with malignant potential---which often appear transiently as the endoscope moves through the lumen. Colonoscopy video analysis aims to identify frames containing visible polyps within the entire video.

\paragraph{Dataset}
Experiments are conducted on a widely used colonoscopy video benchmark integrating the Hyper-Kvasir~\cite{borgli2020hyperkvasir} and LDPolypVideo~\cite{ma2021ldpolypvideo} datasets. The combined dataset contains over one million frames collected from routine screening procedures, covering diverse polyp appearances with varying sizes, shapes, and textures under different illumination and viewing conditions. Following standard protocols~\cite{tian2022contrastive}, the training set includes 61 videos without visible polyps and 102 videos containing polyps, while the test set comprises 30 and 60 videos, respectively. Training videos are annotated at the video level, whereas test videos provide frame-level labels for evaluation.

\paragraph{Metrics}
Following prior studies~\cite{tian2022contrastive}, frame-level results are evaluated using \ac{AUC} and \ac{AP}.

\paragraph{Compared methods}
Both general and domain-specific methods are included for evaluation under the same setting.
\begin{itemize}
    \item \textit{General methods.} Weakly supervised video anomaly detection baselines from natural video domains, including GCN-Ano~\cite{zhong2019graph}, CLAWS~\cite{zaheer2020claws}, MIST~\cite{feng2021mist}, RTFM~\cite{tian2021weakly}, and UR-DMU~\cite{zhou2023dual}, are used for comparison.
    \item \textit{Medical methods.} Weakly supervised polyp detection approaches (\ie, CTMIL~\cite{tian2022contrastive} and Fadmb~\cite{luo2025fadmb}), the endoscopic foundation model Endo-FM~\cite{wang2023foundation}, and EchoPrime~\cite{nature2025}, a large-scale ultrasound model, are included.
    \item \textit{Ours.} Each colonoscopy video is divided into 32 snippets of 16 frames as input, and \acs{ddp} produces snippet-level scores that are propagated to all frames for evaluation.
\end{itemize}

\paragraph{Quantitative results} \cref{tab:polyp_anomaly} summarizes the results on the combined dataset. \model achieves \textbf{94.8}\% \acs{AP}, surpassing the previous best method (\ie, Fadmb~\cite{luo2025fadmb}) by 2.6\%. This performance gain highlights the advantage of counterfactual reasoning in modeling subtle pathological dynamics under weak supervision.

\begin{figure*}[t!]
    \centering
    \small
    \includegraphics[width=\linewidth]{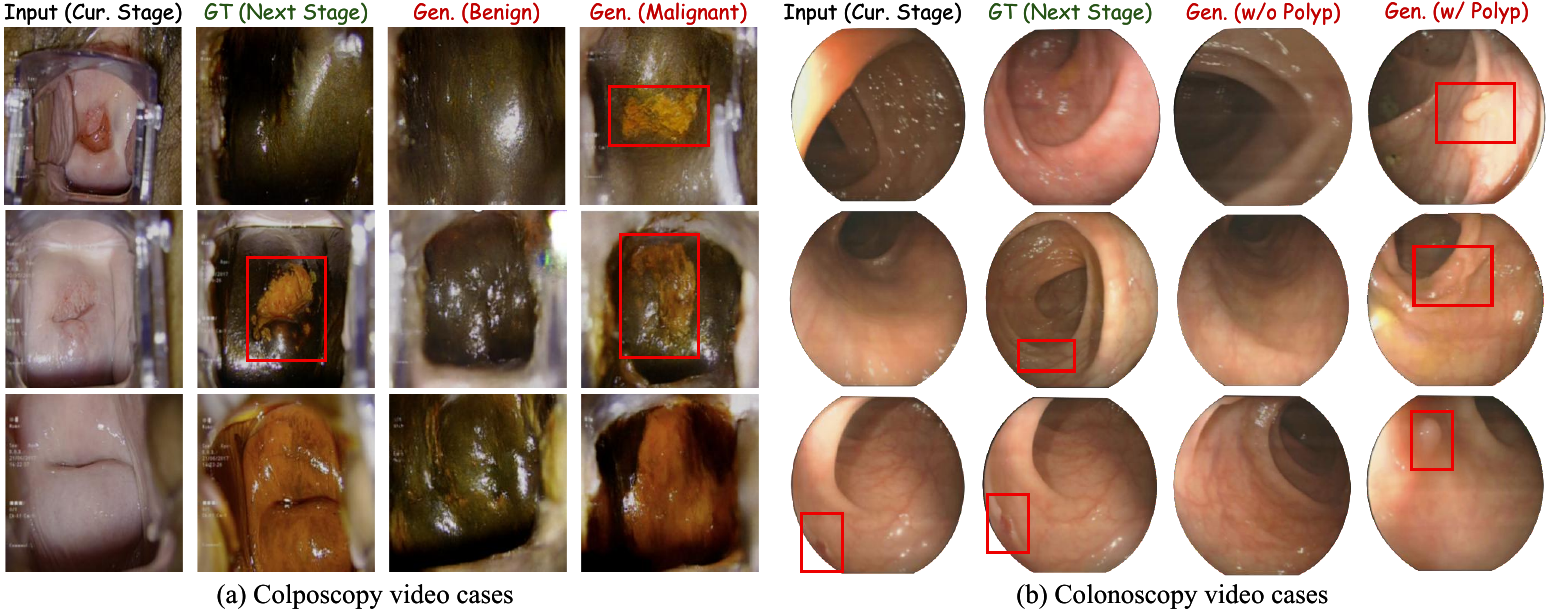}
    \caption{\textbf{Visualization of \acs{cg} (\cref{sec:cg}) across two medical scenarios.} (a)~Colposcopy: synthesizing reagent-induced tissue transitions across examination stages. (b)~Colonoscopy: modeling temporal polyp emergence. Row~1: benign case; Rows~2\&3: malignant cases.}
    \label{fig:visualization}
\end{figure*}

\begin{table}[t!]
    \centering
    \small
    \setlength\tabcolsep{\mytablecolsep}
    \caption{\textbf{Ablation studies on key components under fully supervised (colposcopy) and weakly supervised (colonoscopy) settings.} Results are averaged over five-fold cross-validation. See \cref{sec:ablation} for details.}
    \label{tab:ablation_key}
    \begin{tabular}{ccccc}
        \toprule
        & \multicolumn{2}{c}{Component} & Colposcopy & Colonoscopy \\
        \multirow{-2}{*}{\#} & CRs & DDP & Recall@1$\uparrow$ & AP$\uparrow$ \\
        \midrule
        1 & \xmark & \xmark & 77.9 & 82.8 \\
        2 & \cmark & \xmark & 89.4 & 91.6 \\
        3 & \xmark & \cmark & 80.2 & 85.5 \\
        4 & \cmark & \cmark & \textbf{93.0} & \textbf{94.8} \\
        \bottomrule
    \end{tabular}%
\end{table}

\subsection{Ablation Study and Visualization}\label{sec:ablation}

Comprehensive ablation studies validate the effectiveness of each component across fully supervised (colposcopy) and weakly supervised (colonoscopy) settings.

\paragraph{Key component analysis}
\cref{tab:ablation_key} evaluates the contribution of the clinical rules (CRs, \cref{sec:crl}) and \ac{ddp} strategy (\cref{sec:ddp}) across two medical tasks. (i)~Row~\#1 represents the baseline model containing only the medical video learner (\cref{sec:crl}), reflecting the performance achievable by fine-tuning a pretrained visual encoder (\ie, I3D~\cite{carreira2017quo}). (ii)~Row~\#1 \vs Row~\#2: introducing CRs notably improves performance across both tasks (\ie, $77.9\%\!\rightarrow\!\textbf{89.4}\%$ in Recall@1 for colposcopy), indicating that clinical knowledge injection enhances representation quality. (iii)~Row~\#1 \vs Row~\#3: \acs{ddp} also contributes to improved diagnostic performance (\ie, $82.8\%\!\rightarrow\!\textbf{85.5}\%$ in \acs{AP} for colonoscopy), demonstrating the value of integrating counterfactual reasoning into prediction. (iv)~Row~\#2 \vs Row~\#3: CRs outperform \ac{ddp} across both tasks, particularly in colposcopy (\ie, $80.2\%\!\rightarrow\!\textbf{89.4}\%$ in Recall@1), further underscoring the importance of clinical knowledge for diagnostic representation. (v)~Row~\#4 yields the best overall results, confirming the complementary effects of CRs and \ac{ddp}.

\paragraph{\ac{cg} evaluation (\cref{sec:cg})}
\cref{fig:visualization} presents representative counterfactual generation results across colposcopy and colonoscopy scenarios. For colposcopy (\cref{fig:visualization}(a)), across the acetic-acid to iodine progression, \ac{cg} generates benign and malignant hypotheses for each factual frame. Row~1 shows a benign case where the malignant hypothesis adds subtle yellow--rough textures at clinically plausible locations, revealing early malignant tendencies not visible in the original frame. Row~2 presents a malignant case where the benign hypothesis restores smooth epithelial appearance while preserving lighting and geometry. Row~3 depicts a case with widespread lesions where the benign hypothesis remains coherent and structure-preserving. For colonoscopy (\cref{fig:visualization}(b)), \ac{cg} synthesizes next-stage appearances conditioned on the current observation, generating both \textit{w/o} and \textit{w/} polyp outcomes while preserving temporal coherence and mucosal structure. Notably, Row~3 reconstructs a complete polyp structure from an ulcerated surface, demonstrating anatomical reasoning beyond texture transformation.

\paragraph{\ac{crl} module assessment (\cref{sec:crl})}
We further examine the influence of the \ac{crl} module on diagnostic performance.

\begin{table}[t!]
    \centering
    \small
    \setlength\tabcolsep{\mytablecolsep}
    \caption{\textbf{Effectiveness of individual clinical rules (\cref{sec:crl}):} temporal consistency (Temp.), pathological separability (Sep.), and counterfactual alignment (Align.). Results are obtained through five-fold cross-validation. See \cref{sec:ablation} for details.}
    \begin{tabular}{cccccc}
        \toprule
        & \multicolumn{3}{c}{Clinical Rules} & Colposcopy & Colonoscopy \\
        \multirow{-2}{*}{\#} 
        & Temp. & Sep. & Align. 
        & Recall@1$\uparrow$ & AP$\uparrow$ \\
        \midrule
        1 & \xmark & \xmark & \xmark & 80.2 & 85.5 \\
        2 & \cmark & \xmark & \xmark & 84.1 & 88.0 \\
        3 & \xmark & \cmark & \xmark & 87.5 & 90.2 \\
        4 & \xmark & \xmark & \cmark & 90.8 & 93.0 \\
        5 & \cmark & \cmark & \cmark & \textbf{93.0} & \textbf{94.8} \\
        \bottomrule
    \end{tabular}
    \label{tab:rule_ablation}
\end{table}

\textit{Effectiveness of clinical rules.} We first analyze the contribution of individual clinical rules within the \ac{crl} module. \cref{tab:rule_ablation} shows that all three rules contribute positively to diagnostic performance across tasks. The counterfactual alignment rule provides the largest improvement (Row~\#1 \vs Row~\#4, $80.2\%\!\rightarrow\!\textbf{90.8}\%$ in Recall@1 for colposcopy), demonstrating its critical role in linking factual and counterfactual representations for clinically coherent reasoning.

\begin{table}[t!]
    \centering
    \small
    \setlength\tabcolsep{\mytablecolsep}
    \caption{\textbf{Effectiveness of different visual backbones within the medical video learner (\cref{sec:crl}).} Results are obtained through five-fold cross-validation. See \cref{sec:ablation} for details.}
    \label{tab:encoder_ablation}
    \resizebox{\linewidth}{!}{%
        \begin{tabular}{cccc}
            \toprule
            & & Colposcopy & Colonoscopy \\
            \multirow{-2}{*}{\#} & \multirow{-2}{*}{Backbone} & Recall@1$\uparrow$ & AP$\uparrow$ \\
            \midrule
            1 & Ours (CLIP-ViT/B)\pub{ICML21}~\cite{radford2021learning}          & 84.5 & 89.0 \\
            2 & Ours (DINOv2-ViT/L)\pub{ICCV23}~\cite{oquab2024dinov}            & 86.0 & 90.2 \\
            3 & Ours (ResNet101\pub{CVPR16})~\cite{he2016deep}                    & 89.1 & 91.1 \\
            4 & Ours (R(2+1)D)\pub{CVPR18}~\cite{tran2018closer}                  & 89.8 & 92.0 \\
            5 & Ours (S3D)\pub{ECCV18}~\cite{xie2018rethinking}                   & 92.3 & 94.1 \\
            6 & Ours (I3D)~\pub{CVPR17}~\cite{carreira2017quo}                     & \textbf{93.0} & \textbf{94.8} \\
            \bottomrule
        \end{tabular}%
    }%
\end{table}

\textit{Effectiveness of visual encoder.} We next evaluate the impact of different visual backbones in \cref{tab:encoder_ablation}. Vision-language and self-supervised models show limited transferability to medical domains (\ie, CLIP~\cite{radford2021learning} achieves only 84.5\% Recall@1 in colposcopy). Traditional CNN backbones yield moderate improvements (\ie, ResNet101~\cite{he2016deep} reaches 89.1\% Recall@1). In contrast, video-based architectures exhibit the best overall performance, with I3D~\cite{carreira2017quo} achieving the highest results (\ie, \textbf{93.0}\% Recall@1), underscoring the advantage of spatiotemporal modeling in capturing diagnostic progression.

\section{Conclusion}

This work presents \model, a counterfactual reasoning framework for medical video diagnosis that addresses key limitations of existing approaches. \model integrates three core components: a \ac{cg} that synthesizes realistic tissue transitions under specified pathological states via diffusion modeling, a \ac{crl} module that enforces temporal consistency, pathological separability, and counterfactual alignment through clinically grounded rules, and a \ac{ddp} strategy that combines global temporal context with frame-level counterfactual contrast. Extensive experiments on colposcopy (fully supervised) and colonoscopy (weakly supervised) demonstrate consistent improvements over all baselines, and comprehensive ablation studies verify the contribution of each component. We hope this work encourages further development of clinically grounded, interpretable diagnostic systems for medical video analysis.

{
    \small
    \bibliographystyle{ieeenat_fullname}
    \bibliography{reference_header,references}
}

\clearpage
\appendix
\renewcommand\thefigure{A\arabic{figure}}
\setcounter{figure}{0}
\renewcommand\thetable{A\arabic{table}}
\setcounter{table}{0}
\renewcommand\theequation{A\arabic{equation}}
\setcounter{equation}{0}
\pagenumbering{arabic}
\renewcommand*{\thepage}{A\arabic{page}}
\setcounter{footnote}{0}

\section{Additional Details about Colposcopy}\label{sec:td}

\begin{figure}[b!]
    \centering
    \includegraphics[width=\linewidth]{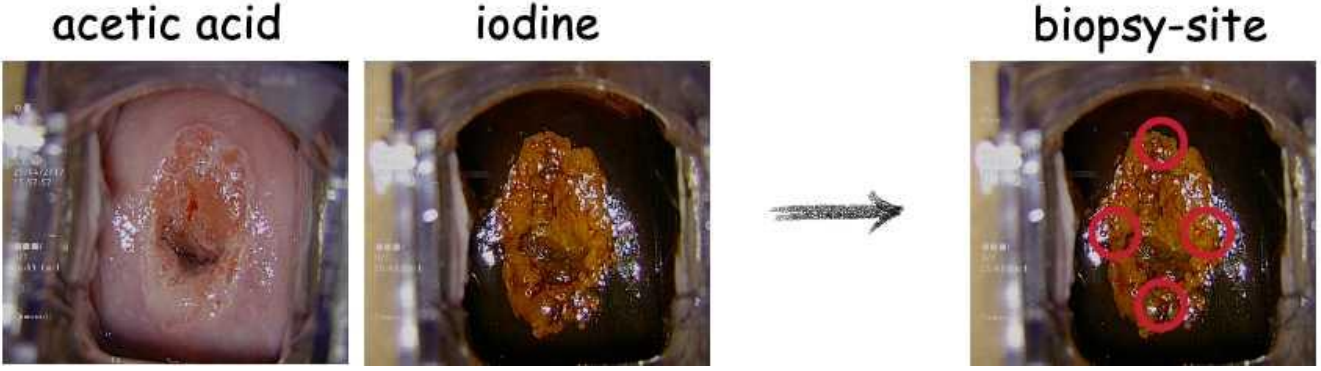}
    \caption{\textbf{Illustration of colposcopic biopsy-site determination.} By comparing tissue responses between the acetic-acid and iodine stages, clinicians localize suspicious areas on the cervix (organized into clock-position regions) for targeted biopsy sampling.}
    \label{fig:addition_colpo}
\end{figure}

\subsection{Task Description}

Colposcopy is a standard procedure for cervical cancer screening, where the cervix is inspected under sequential reagent applications---saline, acetic acid, alcohol, and iodine---to reveal tissue responses indicative of underlying pathology~\cite{schreiberhuber2024cervical}. During the examination, gynecologists determine biopsy sites, \ie, specific anatomical regions from which tissue samples are taken for histopathological confirmation~\cite{xue2025deep}. Clinically, the cervix is divided into \textbf{12 clock-position regions}, and each region may contain one or more suspicious lesions~\cite{hall2023benefits}. Expert colposcopists assess reagent-induced color and texture changes such as acetowhitening, mosaicism, punctation, and iodine-negative areas~\cite{joshi2025randomised}. Among these stages, the transition from acetic-acid reaction to iodine staining provides the most discriminative cues for identifying pathological regions and is therefore the primary basis for biopsy-site localization, as shown in \cref{fig:addition_colpo}. Accordingly, the task is formulated as a multi-label classification problem that predicts biopsy-site locations from the full multi-stage colposcopy video.

\subsection{Training Details}

Each examination contains a four-stage colposcopy video. For temporal modeling, the full sequence is divided into overlapping 16-frame clips using a temporal stride of 8. The overlap preserves smooth temporal continuity across stages and ensures that reagent-induced tissue changes are consistently captured for representation learning.

\section{Additional Experiment}\label{sec:addition_exp}

To assess the generalization ability of \model beyond video-based diagnosis, we evaluate it on a static imaging task, \ie, bilateral mammography classification.

\subsection{Mammography Image Analysis}\label{sec:MIA}

\paragraph{Task description}
Mammography is the primary imaging technique for breast cancer screening and diagnosis~\cite{wang2021bilateral}. This experiment is designed to verify the generalization capability of the proposed counterfactual reasoning framework. The breasts exhibit natural bilateral symmetry, where lesions on one side rarely appear in the corresponding region of the opposite breast~\cite{zhu2017deep}. Therefore, radiologists identify potential malignancies by examining asymmetries between paired views. This task involves detecting pathological lesions based on paired mammograms.

\begin{table}[t!]
    \centering
    \small
    \setlength\tabcolsep{\mytablecolsep}
    \caption{\textbf{Quantitative results of mammography image analysis on INBreast~\cite{moreira2012inbreast}.} Results are obtained through five-fold cross-validation. See \cref{sec:MIA} for details.}
    \label{tab:MIA}
    \begin{tabular}{ccc}
        \toprule
        \multirow{1}{*}{Category} & \multirow{1}{*}{Methods} & AUC $\uparrow$ \\
        \midrule
        \multirow{2}{*}{\textit{General}} 
        & ResNet50\pub{CVPR16}~\cite{he2016deep} & 69.0 \\
        & CNN-based\pub{MICCAI16}~\cite{dhungel2016automated} & 76.0 \\
        \midrule
        \multirow{3}{*}{\textit{Breast-specific}} 
        & Zhu \textit{et al.}\pub{MICCAI17}~\cite{zhu2017deep} & 86.0 \\
        & Wu \textit{et al.}\pub{TMI19}~\cite{wu2019deep} & 86.3 \\
        & Wang \textit{et al.}\pub{TIP21}~\cite{wang2021bilateral} & 90.9 \\
        \midrule
        \multicolumn{2}{c}{\textbf{\textsc{MedVCR} (Ours)}} & \textbf{93.4} \\
        \bottomrule
    \end{tabular}%
\end{table}

\paragraph{Dataset}
Experiments are performed on the public INBreast dataset~\cite{moreira2012inbreast}, a high-quality benchmark for mammogram analysis. The dataset encompasses diverse lesion types, including masses, calcifications, and architectural distortions, representing challenging clinical scenarios. It contains 410 mammograms from 115 cases, each with paired left--right views and image-level BI-RADS annotations verified by pathological examination. Following standard practice~\cite{wang2021bilateral}, images are labeled as malignant if BI-RADS $> 3$, and benign otherwise. For bilateral analysis, 91 valid left--right pairs are used for five-fold cross-validation.

\paragraph{Metrics}
Performance is evaluated using AUC, following~\cite{wang2021bilateral}.

\paragraph{Compared methods}
Bilateral mammogram analysis is formulated as a binary classification task. Evaluation involves three categories of representative methods.
\begin{itemize}
    \item \textit{General methods.} General image classification backbones, \ie, ResNet50~\cite{he2016deep} and CNN-based method~\cite{dhungel2016automated}, are employed as standard discriminative baselines.
    \item \textit{Breast-specific methods.} These methods explicitly model bilateral breast symmetry for classification, including Zhu \etal~\cite{zhu2017deep}, Wu \etal~\cite{wu2019deep}, and Wang \etal~\cite{wang2021bilateral}.
    \item \textit{Ours.} The proposed framework is adapted for mammography by removing $\mathcal{F}^t$ and retaining only $\mathcal{F}^e$, which independently processes each mammogram image.
\end{itemize}

\paragraph{Quantitative results}
Results on INBreast~\cite{moreira2012inbreast} are presented in \cref{tab:MIA}. \model significantly surpasses the previous best method, Wang \etal~\cite{wang2021bilateral} (\eg, $90.9\%\!\rightarrow\!\textbf{93.4}\%$ in AUC). This demonstrates that our framework generalizes effectively to medical image analysis, validating its versatility across different clinical imaging modalities.

\begin{figure}[t!]
    \centering
    \includegraphics[width=\linewidth]{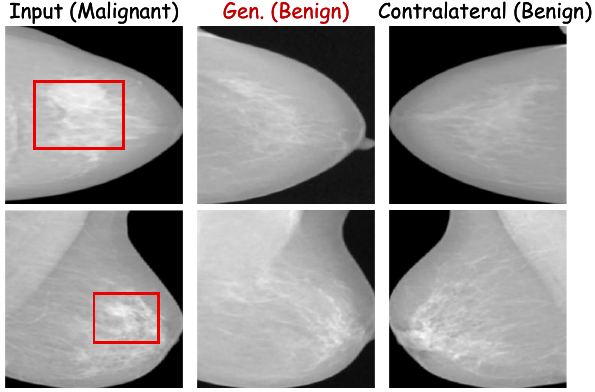}
    \caption{\textbf{Visualization of mammography counterfactuals.} Given a malignant input (left), the model generates a benign hypothesis (middle), which closely aligns with the appearance of the true contralateral breast (right). See \cref{sec:MIA} for details.}
    \label{fig:MIA}
\end{figure}

\paragraph{Qualitative results}
To further examine the behavior of the \acs{cg} on static imaging, \cref{fig:MIA} presents qualitative results on malignant mammograms. For each malignant input (left), the generator produces a benign counterfactual hypothesis (middle). For reference, we also show the true contralateral breast (right), which is typically benign. The generated benign counterfactual suppresses malignant high-density regions and restores coherent glandular and parenchymal patterns. Its morphology resembles the true contralateral breast, demonstrating that our generator preserves global breast architecture while selectively removing pathology-specific structural abnormalities.

\section{Discussion}\label{sec:discussion}

\paragraph{Limitations and future work}
(i)~\textit{Scope of generative modeling.} Our \acs{cg} is trained only to model short-range transitions between adjacent clinical stages. Its ability to synthesize longer-term or cross-stage evolution has not been evaluated and may limit applicability to procedures with complex temporal dynamics.
(ii)~\textit{Coverage of clinical knowledge.} The clinical rules incorporated in \model reflect key diagnostic principles but do not encompass the full range of reasoning strategies used by experienced clinicians. Additional domain knowledge or adaptive rule learning may further improve representation fidelity.
(iii)~\textit{Generality across modalities and institutions.} Although the framework shows strong performance across colposcopy, colonoscopy, and mammography, broader validation on multi-center datasets and diverse imaging modalities is necessary to assess the robustness of \model.

\paragraph{Broader impact}
This work explores the potential of counterfactual reasoning for medical video diagnosis. By generating clinically plausible benign--malignant hypotheses and explicitly modeling temporal tissue transitions, \model provides more transparent and clinically aligned diagnostic cues than conventional end-to-end methods. We hope this work encourages further development of clinically grounded, interpretable diagnostic systems for medical video analysis.

\end{document}